\documentclass[conference]{IEEEtran}

\usepackage{cite}
\usepackage{amsmath,amssymb,amsfonts}
\usepackage{graphicx}
\usepackage[symbol]{footmisc}
\usepackage{textcomp}
\usepackage{gensymb}
\usepackage{xcolor}
\usepackage{longtable}
\usepackage[margin=1in]{geometry} 
\usepackage{booktabs} 
\usepackage{float}
\usepackage{algorithm}
\usepackage{algpseudocode}
\usepackage{amsmath}
\usepackage{dsfont}
\usepackage{xcolor}
\usepackage{graphicx} 
\usepackage{subcaption}
\usepackage{float} 
\usepackage{hyperref}
\def\BibTeX{{\rm B\kern-.05em{\sc i\kern-.025em b}\kern-.08em
    T\kern-.1667em\lower.7ex\hbox{E}\kern-.125emX}}
\begin{document}

\title{RaCIL: Ray Tracing based Multi-UAV Obstacle Avoidance through Composite Imitation Learning
}
\author{\IEEEauthorblockN{Harsh Bansal\textsuperscript{*}}
\and
\IEEEauthorblockN{Vyom Goyal\textsuperscript{*}}
\and
\IEEEauthorblockN{Bhaskar Joshi}
\and
\IEEEauthorblockN{Akhil Gupta}
\and
\IEEEauthorblockN{Harikumar Kandath \textsuperscript{\textdagger}}
}
\maketitle
\footnotetext[1]{These authors contributed equally to this work.}
\footnotetext[2]{Harsh Bansal, Vyom Goyal, Bhaskar Joshi, Akhil Gupta and Harikumar Kandath are with RRC, IIIT Hyderabad \{\textit{harsh.bansal}, \textit{vyom.goyal}, \textit{akhil.g}\}@students.iiit.ac.in, \textit{bhaskar.joshi}@research.iiit.ac.in, \textit{harikumar.k}@iiit.ac.in}

\begin{abstract}
% short intro
In this study, we address the challenge of obstacle avoidance for Unmanned Aerial Vehicles (UAVs) through  an innovative composite imitation learning approach that combines Proximal Policy Optimization (PPO) with Behavior Cloning (BC) and Generative Adversarial Imitation Learning (GAIL), enriched by the integration of ray-tracing techniques. Our research underscores the significant role of ray-tracing in enhancing obstacle detection and avoidance capabilities. Moreover, we demonstrate the effectiveness of incorporating GAIL in coordinating the flight paths of two UAVs, showcasing improved collision avoidance capabilities. Extending our methodology, we apply our combined PPO, BC, GAIL, and ray-tracing framework to scenarios involving four UAVs, illustrating its scalability and adaptability to more complex scenarios. The findings indicate that our approach not only improves the reliability of basic PPO based obstacle avoidance but also paves the way for advanced autonomous UAV operations in crowded or dynamic environments. 
\end{abstract}

\section{Introduction}

The incorporation of Unmanned Aerial Vehicles (UAVs) into both civilian and military activities has shown a significant advancement in the domain, driven by progress in sensor technology, control algorithms, and communication systems. These advancements have made UAVs indispensable for tasks ranging from package delivery to disaster response.The challenge of navigating UAVs through complex environments without encountering obstacles is significant \cite{uav-challenges}, leading to a focus on developing more adaptable navigation solutions beyond traditional methods like Simultaneous Localization and Mapping (SLAM)\cite{slam-uav}.

Recent years have seen the emergence of Reinforcement Learning (RL) as a promising approach for improving UAV autonomy.

\begin{figure}[H]
    \centering
    \includegraphics[width=0.75\linewidth]{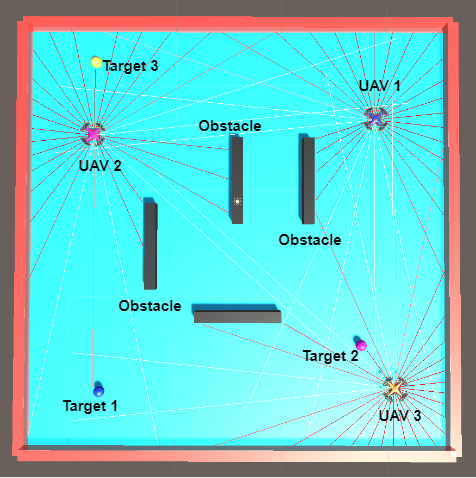}
    \caption{\textbf{UAV Obstacle Avoidance:} A UAV (Blue) aims to reach its goal while navigating around obstacles (Black) and avoiding collisions with other UAVs (Purple, Yellow) in a shared environment.}
    \label{fig:env}
\end{figure}
Alongside RL, Imitation Learning (IL) has shown great potential by learning from expert to navigate similar conditions effectively, streamlining the training process for autonomous control systems.

Deep learning and Generative Adversarial Networks (GANs)\cite{gan}, have significantly enhanced IL,  by generating synthetic but realistic training data that mimics expert behavior, allowing the learning model to improve its predictions on unseen data. IL, in particular, is noted for its efficiency and effectiveness in learning through demonstrations, offering a quicker and more intuitive path to achieving desired behaviors compared to RL's trial-and-error approach. This efficiency, coupled with the interactive nature of IL with human input, underscores its importance in the development of sophisticated artificial agents. 
\\ \\
The contributions from our study are:

\begin{itemize}
    \item It introduces a composite imitation learning framework combining Proximal Policy Optimization (PPO), Behavior Cloning (BC), and Generative Adversarial Imitation Learning (GAIL) with Ray-Tracing techniques for improved UAV Obstacle Avoidance.
    \item Demonstrates the significant impact of integrating Ray-Tracing in enhancing agent training efficiency for detecting and avoiding obstacles.
    \item Showcases the effectiveness of composite IL in coordinating flight paths and enhancing collision avoidance among multiple UAVs when compared to IL solely based on BC.
    \item Validates the model's scalability and robustness through successful applications in environments with increasing UAV numbers, maintaining high success rates showing its potential for advanced autonomous UAV operations.
\end{itemize}

The paper is structured as follows: Section II provides a detailed background, Section III introduces essential preliminaries and the problem formulation, Section IV describes our methodology, Section V examines the results, Section VI concludes with a summary of our findings and contributions to the field and Section VII states the acknowledgements.

\section{BACKGROUND}

 The development of Reinforcement Learning (RL) as an useful technique for way point navigation in intricate environments is a notable milestone in the field of robotics, specifically in the context of Unmanned Aerial Vehicles (UAVs). Reinforcement learning (RL) enables agents to develop optimal behaviour for particular tasks by participating in trial and error experiments with their environment. This process enhances the autonomy of agents in many robotics applications. \\
The study conducted by Almazrouei et al. \cite{background-rl} offers a comprehensive examination of many established techniques and methodologies of the use of reinforcement learning in facilitating the navigation of unmanned aerial vehicles (UAVs) in different scenarios. Zhang et al.\cite{background-dynamic} and Park et al. \cite{background-vision} extend this study to dynamic settings by using a camera as the primary source of Observation and applying various vision based techniques like Google-Net \cite{background-googlenet}, CNN and RNN's to distinguish between stationary and moving objects, hence improving autonomous navigation abilities. \\
To enhance the UAV Navigation across Dynamic Obstacles, Zhang et al.\cite{background-dynamic} also introduces Ray Tracing is a method of encoding or describing the environment. It helps provide useful observations to the agent.\\
The work of Kulic et al. \cite{background-bc} \cite{background-bc2} employ the use of behavioral cloning which tries to learn and imitate the actions of an expert.  \\
Expanding on these approaches, the work of Liu et al.\cite{background-gail}, Fang et al. \cite{background-gail1} \& Shin et al. \cite{background-gail2} use multiple variants of state-of-the-art GAIL to improve and adapt the Agent to randomised dynamic environments. \\
However, these works do not explore the application of Ray Tracing, Behavioral Cloning, GAIL \& Imitation Learning in a Multi-UAV dynamic environment. We thus provide a methodology which helps train an Agent that is both more efficient and adaptable to diverse environments. 

\section{PRELIMINARIES AND PROBLEM FORMULATION}

\subsection{UAV Model}
This study utilizes a simplified UAV model, focusing on two-dimensional movement. The model is defined by position coordinates $[x,\,y]^T$ and controlled through velocity $[v_x,\,v_y]^T$, as described by Eq. (\ref{eq:uav}).

\begin{equation}
    \dot{x} = v_x, \,\,\, 
    \dot{y} = v_y
    \label{eq:uav}
\end{equation}

We assume a constant altitude $z$ for our study in the paper, limiting navigation to the $XY$ plane.

\subsection{Proximal Policy Optimization (PPO)}

Proximal Policy Optimization (PPO) \cite{ppo} is a reinforcement learning method that iteratively updates the policy to maximize reward. PPO maintains stable updates by using a clipped surrogate objective function, which is represented as:

\begin{equation}
\begin{split}
    L_{\text{PPO}}(\theta) = \mathbb{\hat{E}}_t & \left[ \min \left( r_t(\theta) \hat{A}_t, \right. \right. \\
    & \left. \left. \text{clip}(r_t(\theta), 1 - \epsilon, 1 + \epsilon) \hat{A}_t \right) \right]
\end{split}
\label{eq-ppo}
\end{equation}
where $\hat{A}_t$ is the advantage function computed as :
\begin{equation}
    \hat{A}_t = R_{\text{env}} - V(s_t)
\end{equation}
and the ratio $r_t(\theta)$ is defined as:
\begin{equation}
    r_t(\theta) = \frac{\pi_\theta(a_t|s_t)}{\pi_{\theta_{old}}(a_t|s_t)}
\end{equation}

$s_{\text{t}}, a_{\text{t}}$ are state and actions at time t and $\pi_\theta$ represents the policy parameterised on $\theta$. $R_{env}$ is the cumulative reward received from the environment and $V$ denotes the expected cumulative reward.

This ratio is clipped to keep the policy updates within a specified threshold, ensuring small, incremental adjustments to the policy and avoiding large, potentially harmful updates. This balance allows PPO to be both efficient and effective in various learning environments.

\subsection{Imitation Learning}
Imitation Learning is the process of learning from expert demonstrations. Instead of generating random actions and learning through trial and error (as in case of Reinforcement Learning), the agent learns from the expert\cite{il}.

% Imitation Learning creates a mapping between the states and certain actions, represented by $\pi_{\text{imitate}}(s)$. This helps to determine the agent's action in a state $s$. This can be mathematically expressed as: 
% \begin{equation}
%     \pi_{\text{imitate}}(s) = \text{argmax}_{a} P(a|s) 
% \end{equation}

\subsection{Behavioral Cloning}

Behavioral Cloning \cite{bc} \cite{bc2} is a machine learning technique where an agent learns a policy by imitating the behaviour of an expert. It trains a neural network that helps the agent mimic the experts actions.  
\\
Consider a subset $D_{\text{E}}$ consisting of state-action pairs $(s^{E}, a^{E})$, taken  from expert demonstrations over $T$ time steps.
\begin{equation}
    D_{\text{E}} = \{ (s_0^{E}, a_0^{E}), (s_1^{E}, a_1^{E}), \ldots, (s_T^{E}, a_T^{E}) \}
    \label{eq4}
\end{equation}
The behaviour cloning loss function is defined as, a mean squared error (MSE)\cite{bc3}, based on the expert's demonstration data $L_{\text{BC}}$ as:
\begin{equation}
    L_{\text{BC}}(\theta) = \frac{1}{2} \sum_{t}(\pi(s_t^E|\theta) - a_t^{E})^2
\end{equation}
The objective is to minimise the difference between the actions suggested by the policy network $\pi(s_t^E)$, which is characterized by parameters $\theta$, and the expert's chosen actions $a_t^{E}$, corresponding to state $s_t^E$.

\subsection{Generative Adversarial Imitation Learning (GAIL)}
Generative Adversarial Imitation Learning or GAIL \cite{gail}, 
utilizes a discriminator neural network $D_\phi$ to differentiate between the actions of the agent and the expert and rewards the agent if it mimics expert actions. 
\\
$D_E$ contains the state-action pairs generated by the expert demonstrations as defined in \text{eq} \ref{eq4}, and $D_G$ be the state-action pairs generated by the current policy $\pi_\theta$. We update the discriminator parameters with the gradient:
\begin{equation}
\begin{aligned}
 \Bigg( &\mathbb{E}_{\substack{(s,a) \\ \sim D_G}}[\nabla_{\phi}\log D_\phi(s, a)] \\
&+ \mathbb{E}_{\substack{(s,a) \\ \sim D_E}}[\nabla_{\phi}\log(1 - D_\phi(s, a))] \Bigg)
\label{eq6}
\end{aligned}
\end{equation}

The reward generated by GAIL is defined as:
\begin{equation}
        R_{\text{GAIL}} = log(D_{\phi}(s,a))
\end{equation}
and the loss generated for policy updation is calculated as $L_{\text{GAIL}}$ is defined as :
\begin{equation}
 E_{\substack{(s,a) \sim D_E}} \left[\log \pi_{\theta}(a|s) Q(s, a) \right] - \lambda H(\pi_{\theta})
\end{equation}
where
\begin{equation}
Q(\bar{s}, \bar{a}) = \mathbb{E}_{\substack{(s,a) \\ \sim D_E}} \left[ \log(D_{w_{i+1}}(s, a)) \,|\, s_0 = \bar{s}, a_0 = \bar{a} \right]
\end{equation}
$\lambda$ is a control parameter and $H(\pi)$ is the discounted casual entropy which promotes exploration.

\subsection{Problem Formulation}

The objective of the agent is to minimize the distance between the UAV and its corresponding goal $d_{goal}$ as stated in eq. \ref{eq:pf}.
\begin{align}
\label{eq:pf}
\min \quad & d_{\text{goal}}(UAV, goal) \\
\text{subject to} \quad & d_{\text{obstacle}}(UAV, obstacle) \geq \epsilon_{\text{safe}}, \nonumber \\
& \forall \, obstacle \, \in \, Obstacles \nonumber
\end{align}

In our research, we make use of approaches such as Proximal Policy Optimization (PPO), Behaviour Cloning (BC), and Generative Adversarial Imitation Learning (GAIL) with  ray-tracing to enhance UAV's capacity to navigate through the environment in an efficient manner.

\section{Ray Tracing based Composite Imitation Learning (RaCIL)}

\begin{figure}[H]
    \centering
    \includegraphics[width=1\linewidth]{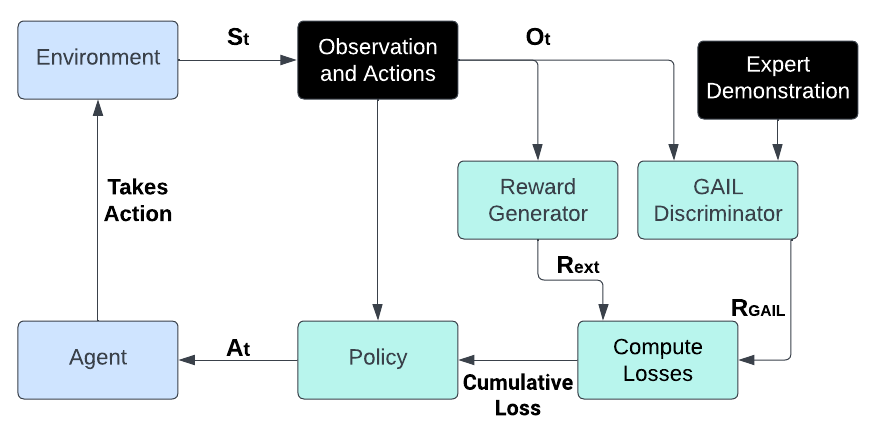}
    \caption{\textbf{System Architecture}: At time $t$ the agent takes action $A\textsubscript{t}$ according to the policy, causing the environment to transition to the next state $S\textsubscript{t}$. Observations from this state ($O\textsubscript{t}$) are sent to the GAIL Discriminator. The reward collected from the environment ($R_{\text{ext}}$) and GAIL ($R_{\text{GAIL}}$) are used to compute the cumulative loss which is used for policy updation.}
    \label{fig:arch}
\end{figure}

Fig. \ref{fig:arch} outlines the sequence of steps undertaken during the model training phase, detailing the process through which the agent selects actions based on a predefined policy. The environment responds by transitioning to a new state and generating relevant observations. These observations are then utilized by both the environment and the GAIL discriminator to generate reward, which is instrumental in the iterative updation and refinement of the policy. This structured approach facilitates a systematic enhancement of the agent's decision-making capabilities. The detailed flow is provided by \textbf{Algorithm 1}.
\\

We first initialise the environment, set our hyper parameters for training and start running PPO algorithm. The model is trained using behaviour cloning for some initial steps. The minimisation of BC loss leads to actions being close to the expert demo. After this we train using GAIL. The total loss is computed by a weighted sum of the losses returned by different algorithms and then we update actor and critic networks on the basis of loss obtained cumulatively from different components of the algorithm.

\begin{algorithm}[H]

\caption{Ray Tracing based Composite Imitation Learning (RaCIL)}
\small
\begin{algorithmic}[1]
\label{algo}
\State Input: Expert Demonstrations \(D_{E}\), training steps \(T\), behaviour-cloning steps \(steps_{\text{bc}}\), RaCIL hyper-parameters ref Table \ref{table:hyperparameters}, Weights for intrinsic and extrinsic rewards \(S_{\text{intrinsic}}\) and \(S_{\text{extrinsic}}\).
\State Output: Trained actor $\pi_{\theta}(a|s)$ and critic networks $Q(s, a|\theta)$ 
\State Initialize the environment for \(N\) UAVs randomly: spawn UAVs, goals, and obstacles, setting up observation space with ray tracing.

\State Randomly initialize actor network $\pi_{\theta}(a|s)$, critic network $Q(s, a|\theta)$  and their weights by $\tilde{\pi}_{\theta'}(a|s)$ and  $Q(s, a|\theta')$ respectively.

\State Initialize empty agent and expert replay buffers $R$ and $R_E$, discriminator \(D_{\phi}\) for GAIL, and cumulative reward \(R\).

\For{BC steps $i = 1, \ldots, steps_{bc}$}
    \State Call TrainUpdate() procedure.
\EndFor
\For{training steps $i = 1, \ldots, T$}
    \State Call TrainUpdate() procedure.
\EndFor

\Procedure{TrainUpdate}{}
\State Observe state \(s\), and decide on action \(a\) using policy \(\pi_{\theta}\).
\State Calculate \(L_{PPO}\) based on the reward received from the environment ref eq. \ref{eq:reward_total}.
\State Initialize \(L_{ext}\) as 0
    \If{Behaviour Cloning is enabled}
        \State Calculate \(L_{BC}\)
        \State Set \(L_{ext}\) = \(L_{BC}\)
        
    \Else
        \State Calculate  \(L_{GAIL}\) 
        \State Set \(L_{ext}\) = \(L_{GAIL}\)
        
    \EndIf
    \State Calculate $L_{cum}$ as \(S_{\text{intrinsic}}\) * \(L_{Env}\) + \(S_{\text{extrinsic}}\) * \(L_{ext}\)
    
    \State Update actor and critic networks according to the computed losses.
    \State Update actor and critic networks on the basis of $L_{cum}$
    % \State $\theta' \leftarrow \tau\theta + (1 - \tau)\theta'$
    % \State $\theta'_Q \leftarrow \tau\theta_Q + (1 - \tau)\theta'_Q$
\EndProcedure
\end{algorithmic}
\end{algorithm}

\subsection{The Environment}

The environment comprises multiple obstacles, multiple UAV agents, and their corresponding goals as shown in Fig. \ref{fig:env}. At the start of each episode, the UAVs are spawned randomly as follows: 
\begin{equation}
\mathbf{X}_{\text{UAV}} = (x_{\text{UAV}}, y_{\text{UAV}}, z_{0})
\end{equation}

where \(x_{\text{UAV}} \sim \text{U}(x_{\text{min}}, x_{\text{max}})\) and \(y_{\text{UAV}}\) is drawn from a combined uniform distribution \(\text{U}(y_{\text{min}}, y_{\text{min}} + r_{\text{min}}) \cup \text{U}(y_{\text{max}} - r_{\text{max}}, y_{\text{max}})\), and \(z_{0}\) represents a fixed altitude.

Subsequently, the goal for the UAV is defined by 
\begin{equation}
    \mathbf{X}_{\text{goal}} = \left(  x_{\text{goal}},  y_{\text{goal}}, z_{0} \right)
\end{equation}
where \(x_{\text{goal}} \sim \text{U}(x_{\text{min}}, x_{\text{max}})\). The \(y_{\text{goal}}\) value depends on \(y_{\text{UAV}}\)'s position: for \(y_{\text{UAV}}\) within \([y_{\min}, y_{\min} + r_{\text{min}}]\), \(y_{\text{goal}}\) is drawn from \(\text{U}(y_{\max} - r_{\text{max}}, y_{\max})\); conversely, if \(y_{\text{UAV}}\) falls in \([y_{\max} - r_{\text{max}}, y_{\max}]\), then \(y_{\text{goal}}\) comes from \(\text{U}(y_{\min}, y_{\min} + r_{\text{min}})\). Both UAVs and goals share the same fixed altitude \(z_{0}\).

Finally, the environment randomly generates a series of obstacles. Let there be $i$ obstacles ($i\in[1,N_{obs}]$), then for the i\textsuperscript{th} obstacle:
\begin{equation}
\mathbf{X}_{\text{obstacle}} = \left( x_{\text{obstacle}}, y_{\text{obstacle}}, z_{0} \right)
\end{equation}

where \(y_{\text{obstacle}} \sim \text{U}(y_{\text{min}} + r_{\text{min}}, y_{\text{max}} - r_{\text{max}})\). The \(x_{\text{obstacle}}\) coordinate is distributed uniformly within \([x_{\text{min}} + d \cdot (i - 1), x_{\text{min}} + d \cdot i]\), where \(d = \frac{x_{\text{max}} - x_{\text{min}}}{N_{obs}}\). Each obstacle is also given a random rotation around the z-axis ranging from 0 to 180 degrees.

Here, $m_{\min}$ and $m_{\max}$ specify the environment dimensions along the $m$-axis ($m \in \{x, y\}$). 
For the agent, the obstacles spawned in the environment are static whereas the other UAV's navigating the environment in a Multi-UAV system act as dynamic obstacles for the agent.\\

\subsection{Observation Space}

We define the observation space as Eq. \ref{obs-space}:
\begin{equation}
\begin{aligned}
\text{Observation Space} = \{ & (x_a, y_a), 
                                (x_g, y_g), \\
                               & \text{Ray Tracing Observations} \}
\end{aligned}
\label{obs-space}
\end{equation}

Where,  $x_a$ and $y_a$ represents the UAV's (agent's) position within the simulation environment along $x$-axis and $y$-axis respectively. $(x_g, y_g)$ represents the goal's position along the $x$-axis and $y_g$ respectively.\\
A ray perception sensor is added to our agent UAV which projects rays in different directions and collects the observations when the rays strike surface. Ray Tracing observations correspond to the observations about the environment collected by this sensor and it includes observations of ray perception sensor include the distance, surface normal, tag and information about the collider of the object hit.
\subsection{Action Space}
The action space for the agent for our study is defined as a discrete set comprising three actions denoted as Move Forward (Fwd), Rotate Left ($\theta_{left}$) and Rotate Right ($\theta_{right}$) :
\begin{equation}
\label{action-space}
\begin{aligned}
    \text{Action Space} = \{ & \text{Fwd}, 
    &  \textbf{$\theta_{left}$}, & 
    &  \textbf{$\theta_{right}$} \}
\end{aligned}
\end{equation}
These actions are carefully chosen to allow the agent to perform comprehensive navigation and precise orientation within the two-dimensional operational plane. The least value by which it moves forward is by 0.04 units and rotates with $2^{\circ}$ clockwise or anti-clockwise.
\subsection{Reward Function} 
Designing effective reward functions is a critical challenge in Reinforcement Learning (RL), often requiring hand-engineering and domain-specific knowledge. Both Imitation Learning (IL) \cite{gail} and Inverse Reinforcement Learning (IRL) \cite{irl} attempt to address this by learning from expert behavior, but they face issues with computational efficiency and the complexity of accurately capturing expert trajectories \cite{challenges-irl}. 

We defined the reward function as a sum of several components, each contributing to the overall reward. For our work, it is expressed as:

\begin{equation}
\begin{aligned}
R_{\text{extrinsic}} =\ & R_{\text{collision}} + R_{\text{proximity}} + R_{\text{time penalty}}
\end{aligned}
\label{eq:reward_total}
\end{equation}

\begin{equation}
\begin{aligned}
R_{\text{collision}} =\ & \begin{cases}
+r_{f} & \text{for reaching its goal}, \\
-r_{f} & \text{for reaching other UAV's goal}, \\
-r_{f} & \text{for collision with other UAV}, \\
-\frac{r_{f}}{2} & \text{for collision with obstacle}.
\end{cases}
\end{aligned}
\end{equation}
\\
\begin{equation}
\begin{aligned}
R_{\text{proximity}} =\ & \begin{cases}
+r_{p} & \text{within $\varepsilon$-proximity of its goal}, \\
-r_{p} & \text{within $\varepsilon$-proximity to other UAV}.
\end{cases}
\end{aligned}
\end{equation}
\\
\begin{equation}
\begin{aligned}
R_{\text{time penalty}} = -r_{tp} \text{ for each time step.}
\end{aligned}
\end{equation}
\subsection{The Agent}
We have used Proximal Policy Optimization (PPO) eq. (\ref{eq-ppo}) for obstacle avoidance. The PPO algorithm consists of both actor and critic components. The actor predicts mean values for each action scalar based on environment observations, while the critic evaluates the perceived value associated with the input state. We are also using Behaviour cloning eq. (\ref{eq4}) and GAIL eq. (\ref{eq6}) by providing expert demonstration. This helps the agent imitate the expert thus leading to faster convergence of the loss function and thus a shorter training time.
\subsection{Ray Tracing}
\begin{figure}
    \centering
    \includegraphics[width=1\linewidth]{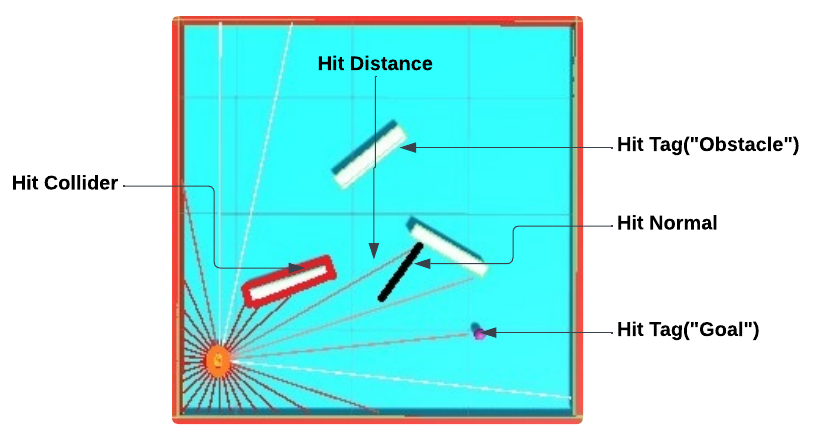}
    \caption{\textbf{Ray Tracing Scenario}: This figure depicts the collection of observation by a UAV using RayTracing}
    \label{fig:rt}
\end{figure}
We have used Ray Tracing in our work, leading to  better navigation and higher accuracy of our model. The ray perception sensor attached to the agent subtends rays with the ray equation:
\begin{equation}
    \begin{aligned}
        \vec{P}(\alpha) = O_{r} + \alpha \cdot \vec{D}
    \end{aligned}
\end{equation}
It provides crucial observations such as { Hit Distance, Hit Tag, Hit Normal, Hit Collider}.
Hit Distance is evaluated as $ \text{hit distance} = \| C_{r}- O_{r} \| $. ($O_{r}$ is the origin point for the ray, $\vec{D}$ is direction vector parameterised on $\alpha$ and $C_{r}$ is the point of collision with the object collider). We have allotted different tags to different types of objects (e.g., "Obstacle" tag or "Goal" tag) so that the agent can identify these and act accordingly.

\section{Results}

This section covers the experimental results obtained during training and evaluating the model in the Unity simulation environment \cite{unity} using Unity ML Toolkit \cite{ml-agents}. The policy training took place on Intel i7 processor and RTX 3050 Ti GPU taking around 5 hours for 10 million time-steps.

\begin{table}[H]
\centering
\caption{{Hyper-parameters used for Training}}
\begin{tabular}{lc|lc}
\toprule
\textbf{Variable Name} & \textbf{Value} & \textbf{Variable Name} & \textbf{Value} \\
\midrule
batch\_size & 1024 & normalize & false \\
buffer\_size & 2048 & hidden\_units & 1024 \\
learning\_rate & 3.0e-4 & num\_layers & 8 \\
beta & 0.005 & extrinsic gamma & 0.99 \\
epsilon & 0.2 & extrinsic strength & 1.0 \\
lambda & 0.95 & gail gamma & 0.99 \\
num\_epoch & 3 & gail\_strength & 1.0 \\
learning\_rate\_schedule & linear &  bc\_strength & 0.5\\
beta\_schedule & constant &  $steps_{\text{bc}}$ & 100000\\
$r_f$ & 10000 & $r_p$ & 0.2 \\
$r_{\text{tp}}$ & 1 & $x_{\text{max}}$ & 15 \\
$y_{\text{max}}$ & 15 & $x_{\text{min}}$ & -15 \\
$y_{\text{min}}$ & -15 & $r_{\text{min}}$ & 3.5 \\
$r_{\text{max}}$ & 3.5 & $i$ (No. of Obstacles) & 4\\
$\varepsilon-\text{proximity}$ & 5\\
\bottomrule
\end{tabular}
\label{table:hyperparameters}
\end{table}

\subsection{Training}
This section explains the Training Phase and the results obtained. The study was initiated with a single UAV (ref Fig. \ref{fig:env} where only UAV1 and target1 are spawned along with the obstacles), and various approaches were attempted, analyzing the pros and cons of Ray Tracing, Behavioral Cloning, and GAIL. Following a comparative analysis, the best-performing combination, i.e., Ray Tracing + Behavioral Cloning + GAIL, was selected. This methodology was then extended to Multi-UAV Environments.
In order to effectively understand the results of the Training Phase we have divided the Training Phase into three separate studies that were made. The individual studies are defined below:\\
\\
\textbf{Study 1: The Critical Role of Ray-Tracing in Observation Spaces}: 
This study highlights the importance of Ray Tracing as a part of the Observation Space for the Agent. 

    \begin{figure}[H]
    \centering
    \subfloat[Mean Reward Progression]{\includegraphics[width=8cm]{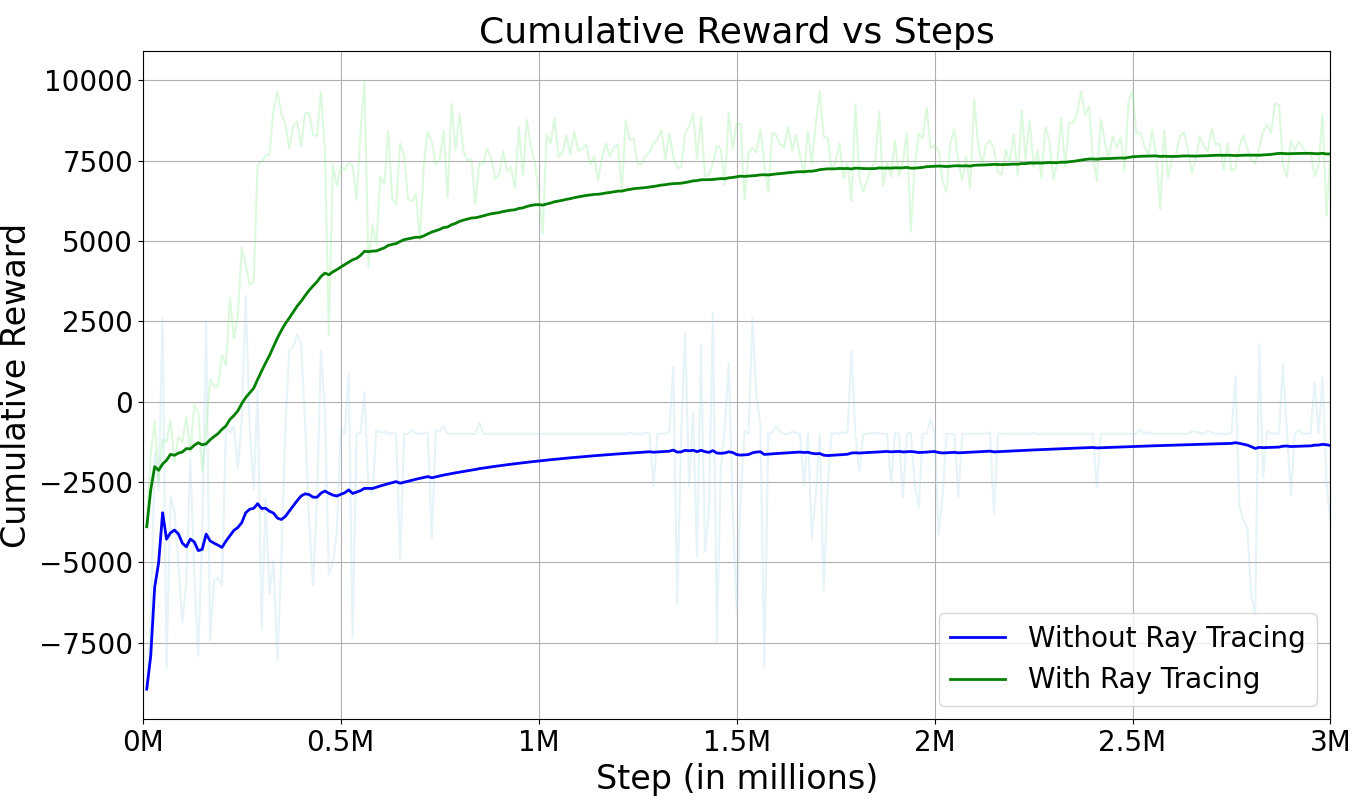} \label{fig:mean-reward}}
    \qquad
    \subfloat[Episode Length Progression]{\includegraphics[width=8cm]{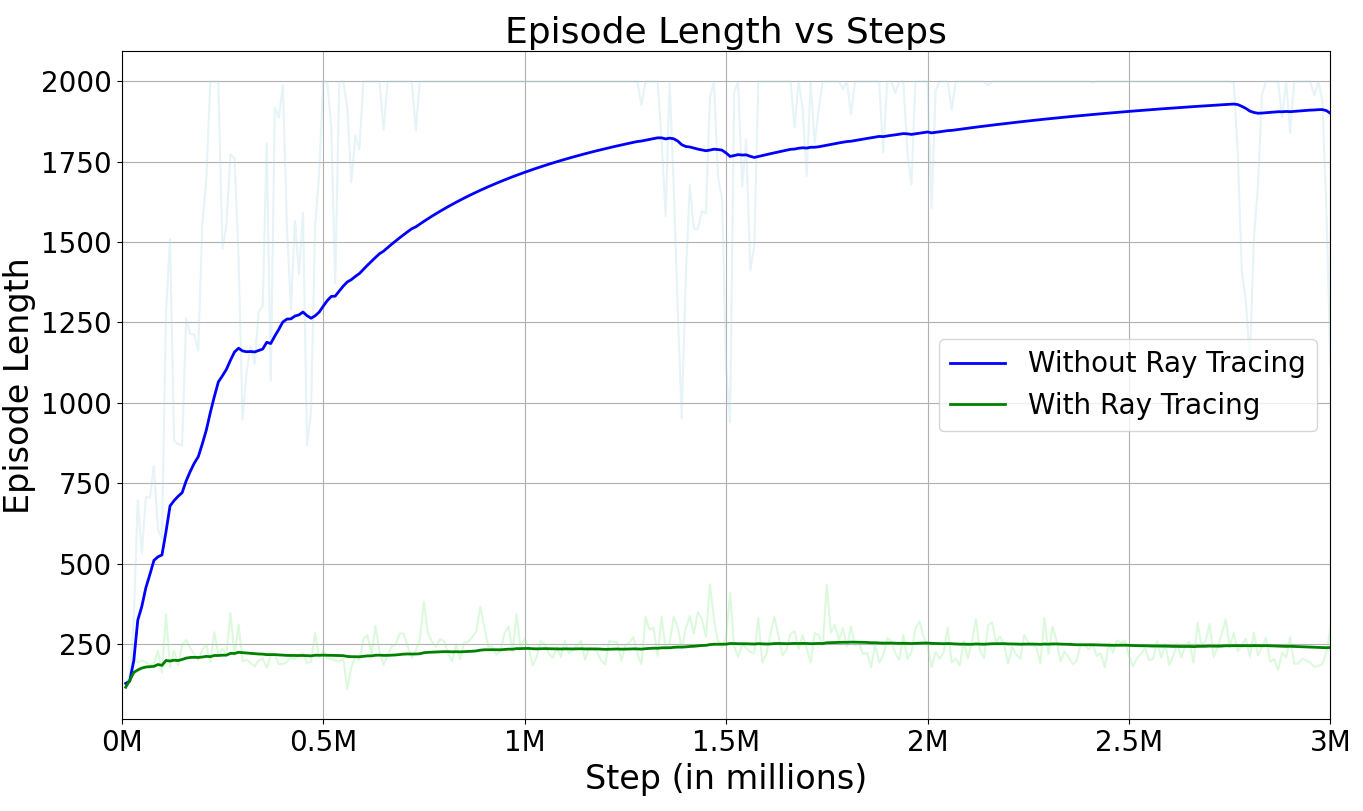} \label{fig:episode-length}}
    \caption{\textbf{Training Results for Study 1:} Performance comparison of UAV navigation with and without Ray Tracing. Fig. \ref{fig:train-study1}(a) illustrates the progression of mean rewards throughout the training process, highlighting the effectiveness of Ray Tracing in enhancing navigational decisions. Fig. \ref{fig:train-study1}(b) shows the episode length progression, indicating improved efficiency and learning speed when Ray Tracing is utilized. }
    \label{fig:train-study1}
\end{figure}

    For this study we have used the environment with a single UAV, goal and multiple obstacles (ref Fig. \ref{fig:env} where only UAV-1 and Target-1 are spawned along with the obstacles) and trained it to reach the goal. In the case of UAV navigation without ray tracing, the model receives observations in the form of location coordinates of the agent, goal, and obstacles. Plots in Fig. \ref{fig:train-study1} (a) and (b) corresponds to Cumulative Reward Vs Steps and Episodic Length Vs Steps. Both the plots of Fig. \ref{fig:train-study1} suggests that utilizing ray tracing for even single UAV environment results in higher cumulative rewards and shorter episode lengths, implying that the ray tracing contributes to a more efficient training.\\
    
 \textbf{Study 2: The Impact of GAIL on Agent Training and Performance}:
 This study focuses on assessing the significance of incorporating GAIL alongside BC.
    The environment for this consists of a UAV navigating through numerous obstacles with the objective of reaching a designated goal. 

    \begin{figure}[H]
    \centering
    \subfloat[Mean Reward Progression with BC vs. BC + GAIL]{\includegraphics[width=8cm]{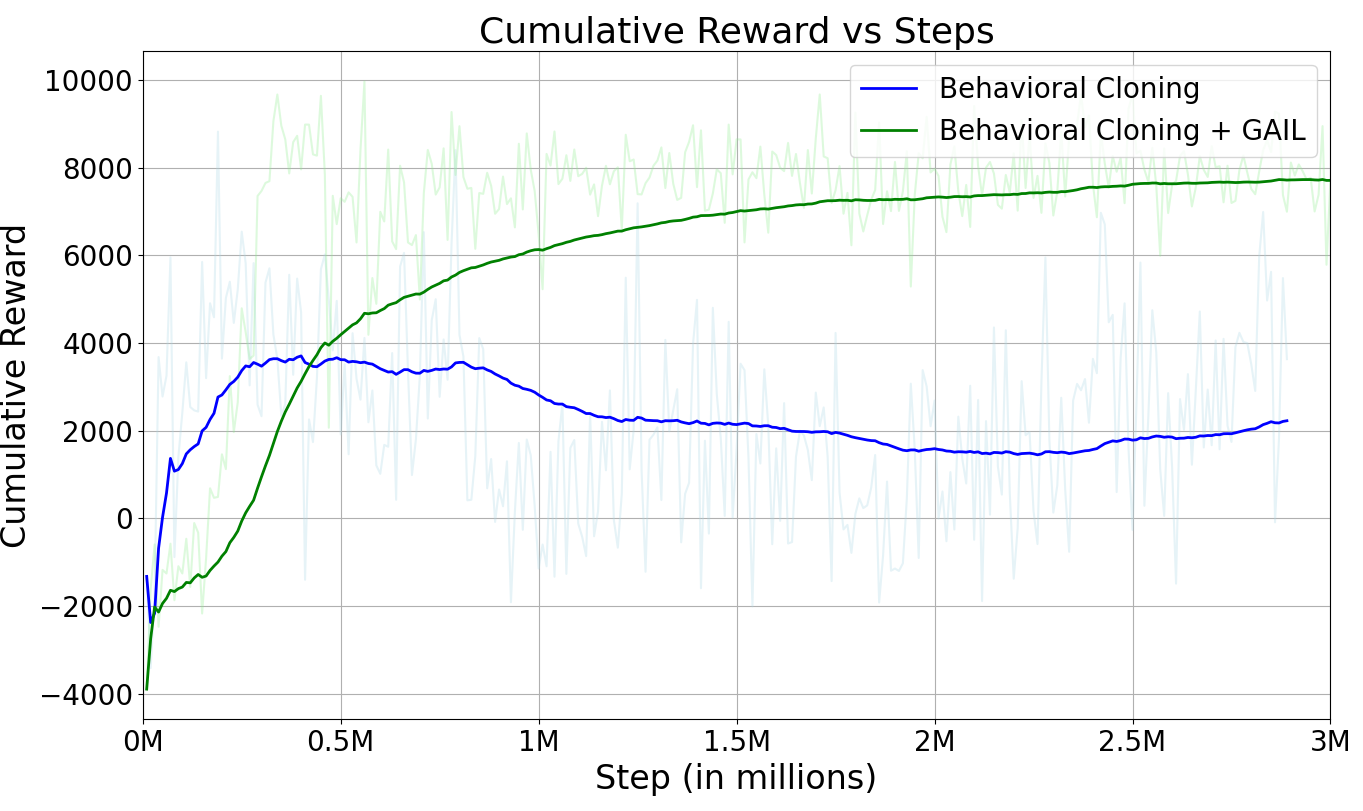} \label{fig:reward-bc-gail}}%
    \qquad
    \subfloat[Episode Length Progression with BC vs. BC + GAIL]{\includegraphics[width=8cm]{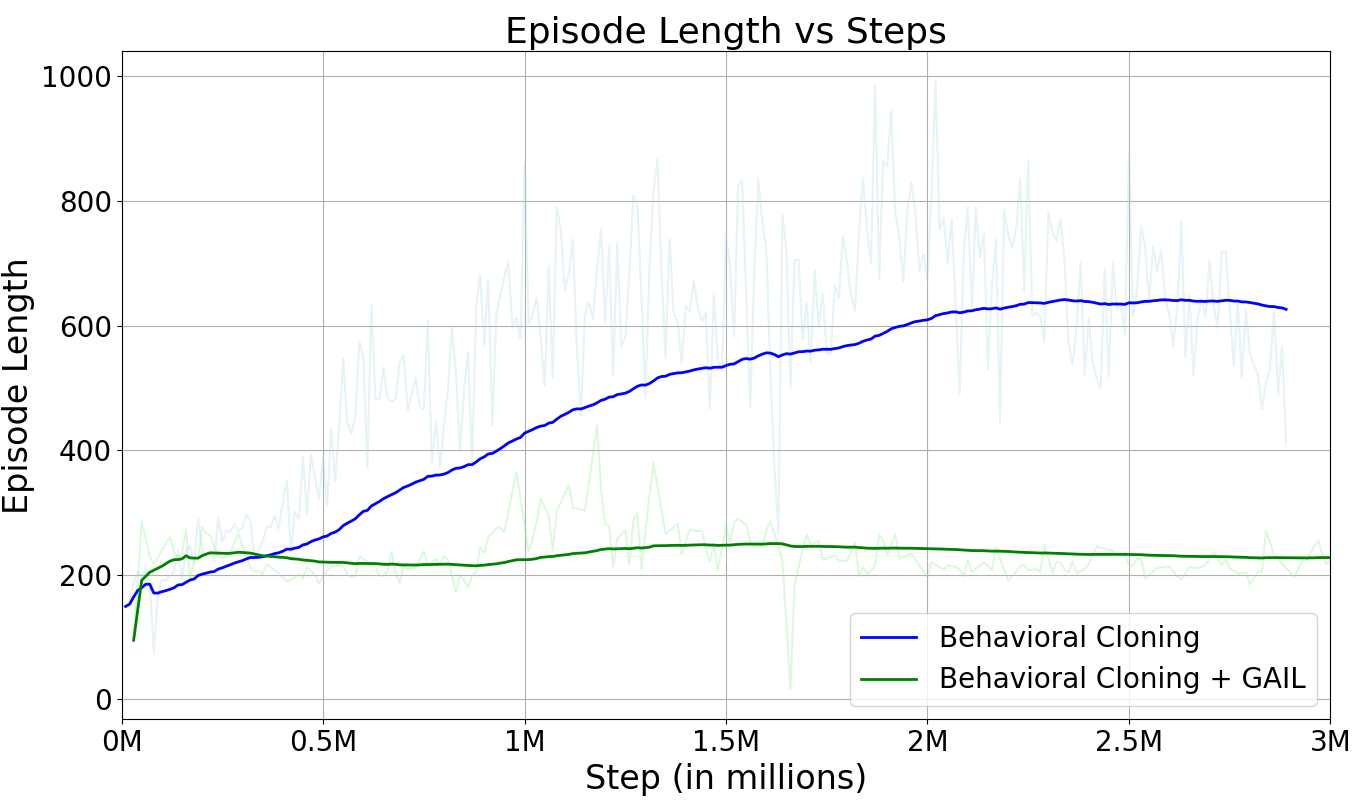} \label{fig:length-bc-gail}}%
    \caption{\textbf{Policy Training Results for Study 2:} Comparative analysis of UAV policy training using Behavior Cloning (BC) and BC integrated with Generative Adversarial Imitation Learning (GAIL). Fig. \ref{fig:train-study2}(a) showcases the mean reward progression, comparing the efficacy of BC alone versus the enhanced approach incorporating BC + GAIL. Fig. \ref{fig:train-study2}(b) illustrates the episode length progression, further demonstrating the impact of GAIL in refining learning efficiency.} 
    \label{fig:train-study2}
\end{figure}

    The performance of two different training approaches is compared: one agent is trained using PPO + BC + Ray Tracing, while the other utilizes PPO + BC + GAIL + Ray Tracing as seen in Fig. \ref{fig:train-study2}. Evaluation of these trained policies reveal that the inclusion     
     of GAIL in the training protocol results in approximately a 17\% improvement in the success rate of the resulting policy as seen in Table \ref{tab:table2}.
\\
\\
\textbf{Study 3: Assessing Scalability and Robustness in UAV Training Across Multiple Agents}: This investigation explores the adaptability and reliability of the agent training framework when scaled from single UAV to 2-UAV and 3-UAV scenario(ref Fig. \ref{fig:env} for the Environment Initialisation). The agent, trained with a single UAV is extended to 2 UAV in  PPO + BC + GAIL + Ray Tracing setting, and that could be further extended to a more congested 3-UAV setting, with the intention of evaluating performance amidst greater obstacle density, including both static obstacles and dynamic obstacles like additional UAVs. The UAVs are uncoupled during the training phase and act according to a centralized policy. Only the individual losses are considered for training the agent policy. The findings revealed that the agent's performance in the 3-UAV scenario was comparable to that in 1-UAV and 2-UAV setting, demonstrating that the agent maintained its effectiveness despite the heightened intricacy of the environment.

This study evaluates the agent's behavior on the following metrics:

\begin{itemize}
    \item \textbf{Mean Reward}: Mean Reward refers to the Cumulative Reward given to the Agent based on a Terminal Condition. Figs. \ref{fig:train-study1}(a), \ref{fig:train-study2}(a) and \ref{fig:train-study3}(a) refers to the Mean Reward Achieved by the Agent as the Number of Training Steps of the Agent increase in the Training Phase.

\begin{figure}
    \centering
    \subfloat[Mean Reward Progression: 2 UAVs vs. 3 UAVs]{\includegraphics[width=8cm]{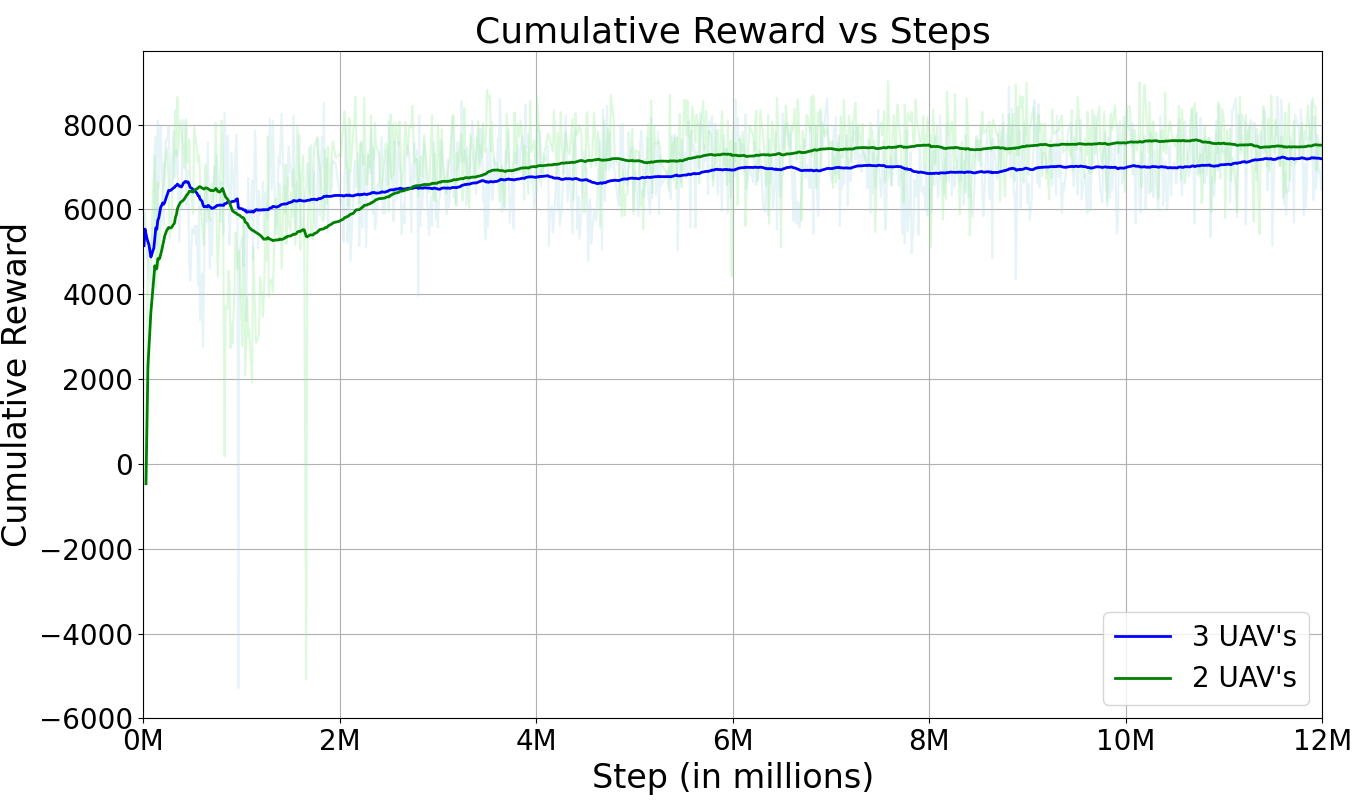} \label{fig:mean-reward-2-3-uavs}}%
    \qquad
    \subfloat[Episode Length Progression: 2 UAVs vs. 3 UAVs]{\includegraphics[width=8cm]{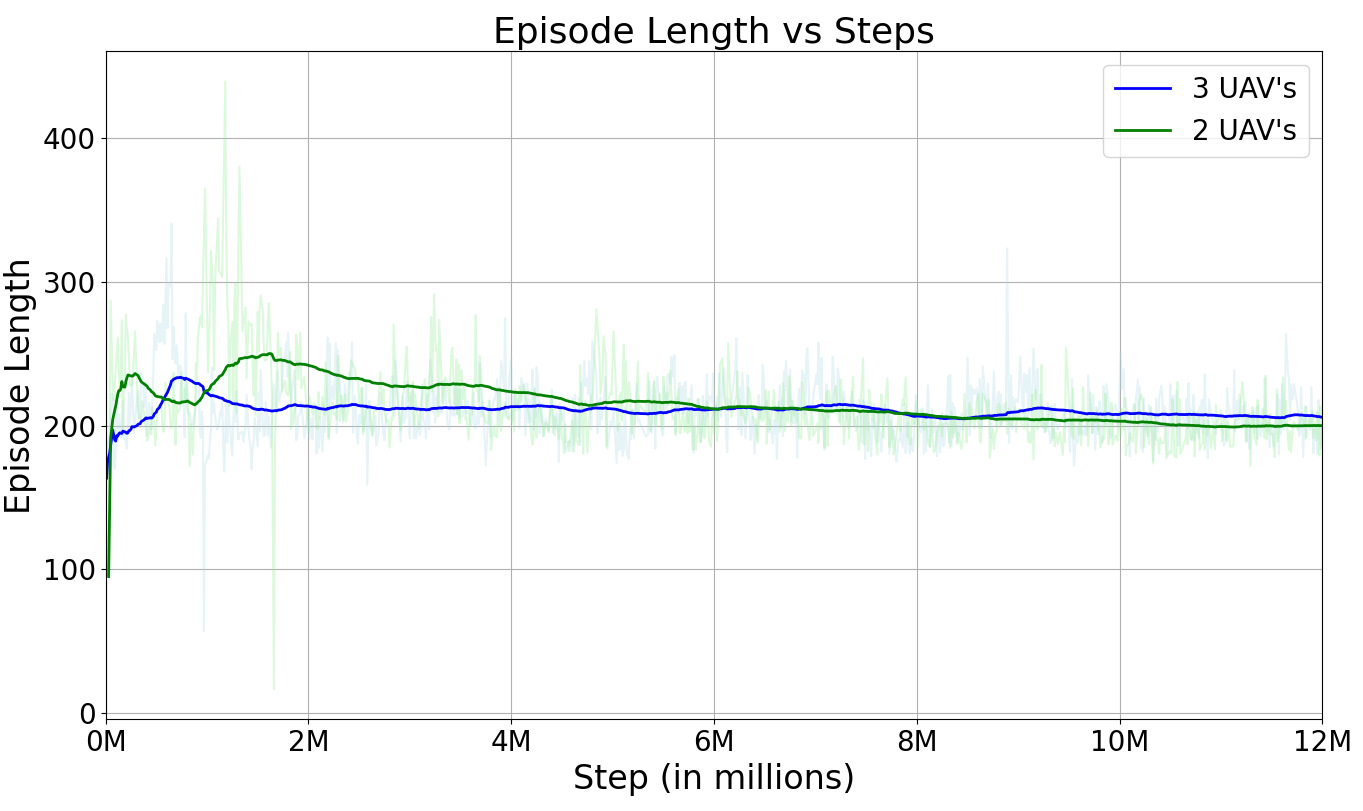} \label{fig:episode-length-2-3-uavs}}%
    \qquad
    \subfloat[GAIL Loss Progression: 2 UAVs vs. 3 UAVs]{\includegraphics[width=8cm]{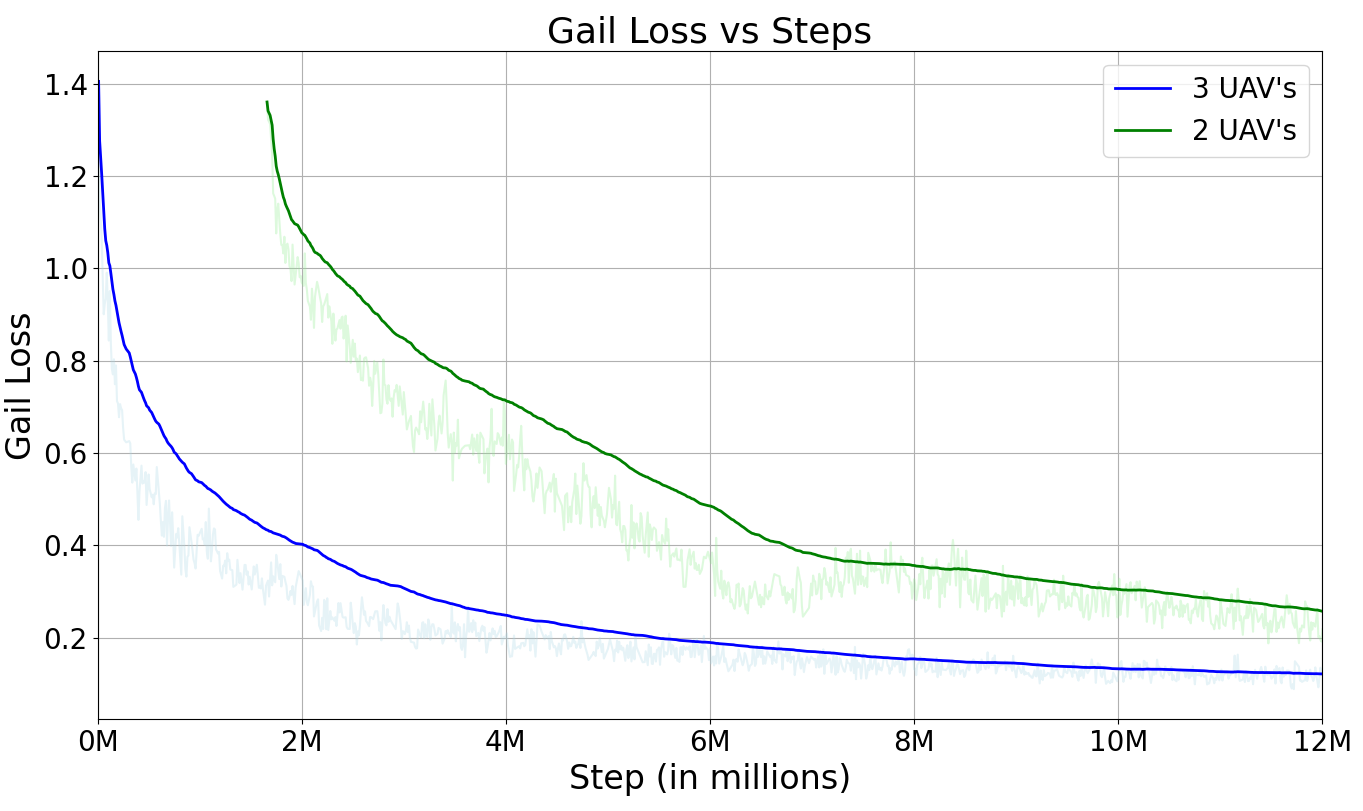} \label{fig:gail-loss-2-3-uavs}}%
    \caption{\textbf{Evaluating Scalability in UAV Obstacle Avoidance Training:} Transitioning from a single UAV to a 2-UAV environment and a 2-UAV environment to a 3-UAV environment. Fig. \ref{fig:train-study3}(a) 
    depicts the progression of mean rewards, illustrating the impact of adding an additional UAV on the training efficiency and overall success. Fig. \ref{fig:train-study3}(b) shows the episode length progression, providing insight into the complexity and challenges introduced with more UAVs. 
    Finally, Fig. \ref{fig:train-study3}(c) presents the GAIL Loss progression, highlighting the learning dynamics and adaptation required in more crowded scenarios.}
    \label{fig:train-study3}
\end{figure}

\footnote{The code for training and evaluation can be found at \href{https://github.com/HarsH-BansaL16/RaCIL}{GitHub}  and the demo video could be found at \href{https://youtu.be/gqLTXekQXmU}{Demo Video} }

    \item \textbf{Episode Length}: Episode Length refers to the Number of Time-steps taken by the Agent such that it reaches a Terminal Condition. Figs. \ref{fig:train-study1}(b), \ref{fig:train-study2}(b) and \ref{fig:train-study3}(b) refers to the Mean Episode Ending Length by the Agent. 
    \item \textbf{GAIL Loss}: GAIL Loss refers to Value Loss of the GAIL Discriminator. Fig. \ref{fig:train-study3}(c), shows the Loss Value that is decreasing with time. 
\end{itemize}

The findings from our study indicate that relying solely on Behavioral Cloning (BC) is insufficient for achieving optimal pathfinding with unmanned aerial vehicles (UAVs) in novel scenarios. The absence of ray tracing further complicates the UAV's ability to learn effective path navigation, especially in environments with dynamic obstacles, hindering the agent's development of a robust avoidance strategy. Additionally, the model demonstrates adaptability with an increase in the number of UAVs, as evidenced in \textbf{Table} \ref{tab:table3}, suggesting its potential scalability and effectiveness in more complex operational contexts.

% Based on the results obtained we can infer that BC alone is not optimal as in this case the UAV is unable to find optimal path in unseen situations. Also without the inclusion of ray tracing the UAV is unable to learn optimal paths as the obstacles are dynamic and thus the agent is unable to learn the policy to avoid them. The model could also adapt well in case the number of UAVs is increased as can be seen in \textbf{Table} \ref{tab:table3}.

% In case of only BC we could see a drastic drop in the Success Rate. This is because in behavioral cloning, the agent performs good on only the environments it has previously seen as a part of the Expert Demonstrations. Whereas in case of BC + GAIL, we could observe that the agent learns to navigate in unseen environment very efficiently with success rate of 92\%. 
% The Success Rate drops in case of 3 UAV system due to the cluterness of the Environment, with 3 UAVs Moving constantly along with 3 Goals and Multiple Obstacles. This leads to the Success Rate dropping from 93\% to 85\% which is still a very great success rate. 

\begin{table}[H]
    \centering
    \caption{Simulated Evaluations of Trained Policies for Different Cases}
    \label{tab:table2}
    \begin{tabular}{p{0.2\columnwidth} p{0.4\columnwidth} p{0.2\columnwidth}}
        \toprule
        \textbf{Environment} & \textbf{Agent Policies} & \textbf{Success Rate} \\
        \midrule
        1 UAV  & PPO + BC  & 24\% \\
        1 UAV  & PPO + BC + Ray Tracing & 76\%\\
        1 UAV  & PPO + BC + GAIL + Ray Tracing & 93\% \\
        \bottomrule
    \end{tabular}
\end{table}

\begin{table}[H]
    \centering
    \caption{Simulated Evaluations of Trained Policies for Scalability and Robustness}
    \label{tab:table3}
    \begin{tabular}{p{0.2\columnwidth} p{0.4\columnwidth} p{0.2\columnwidth}}
        \toprule
        \textbf{Environment} & \textbf{Agent Policies} & \textbf{Success Rate} \\
        \midrule
        2 UAVs  & PPO + BC + GAIL + Ray Tracing & 92\% \\
        3 UAVs  & PPO + BC + GAIL + Ray Tracing & 85\% \\
        \bottomrule
    \end{tabular}
\end{table}

\section{Conclusion}

The proposed research has demonstrated the effectiveness of utilizing a composite imitation learning technique within the Unity simulation environment to enhance UAV autonomy and obstacle avoidance capabilities. Through the application of ray-tracing with a Behavior-Cloning neural network architecture and the exploration of Generative Adversarial Imitation Learning (GAIL), we have shown promising results in training agents to navigate around obstacles and reach predefined goal locations. The addition of GAIL and ray-tracing significantly improved performance compared to the baseline model, which used the PPO algorithm for training the agent along with behavioral cloning on human expert demonstrations.

By leveraging Unity as the simulation platform, we have conducted a systematic exploration of imitation learning techniques, contributing to the advancement of UAV navigation methodologies in dynamic and challenging environments as well as showing the scalability of our approach to multi-UAV navigation scenarios. Our findings have implications for real-world applications requiring reliable and adaptive navigation strategies, paving the way for future research in this domain.
In future, we plan to scale the model to 3D and deploy it to real drones.

\section{Acknowledgement}
The authors acknowledge the support provided by MeitY, Govt. of India, under the project ”Capacity Building for
Human Resource Development in Unmanned Aircraft System (Drone and related Technology)”

\end{document}